\documentclass{article} 
\usepackage{nips14submit_e,times}
\usepackage{hyperref}
\usepackage{url}

\usepackage{graphicx} 
\usepackage{amssymb} 
\usepackage{amsmath}
\usepackage{bm} 
\usepackage{caption}
\usepackage{subcaption}
\usepackage{multirow}
\usepackage{tikz}
\usetikzlibrary{shapes.gates.logic.US,trees,positioning,arrows,shapes.geometric}
\tikzset{
    factor/.style={
        draw,
        shape border rotate=90,
        isosceles triangle,
        isosceles triangle apex angle=60,
        node distance=3em,
        minimum height=2em
    },
    block/.style={
        rectangle,
        draw, 
        minimum width=4em,
        minimum height=1.8em
    },    
    sblock/.style={
        rectangle,
        draw, 
        minimum width=2em,
        minimum height=1.8em
    },
}

\DeclareMathOperator*{\argmin}{arg\,min}

\title{``Mental Rotation'' by Optimizing \\ Transforming Distance}

\author{
Weiguang Ding\\
School of Engineering\\
University of Guelph\\
Guelph, Ontario, Canada\\
\texttt{wding@uoguelph.ca} \\
\And
Graham W. Taylor \\
School of Engineering\\
University of Guelph \\
Guelph, Ontario, Canada\\
\texttt{gwtaylor@uoguelph.ca} \\
}

\nipsfinalcopy 

\begin{document}

\maketitle
\vspace{-.75cm} 
\begin{abstract}
  The human visual system is able to recognize objects despite
  transformations that can drastically alter their
  appearance. To this end, much effort has been devoted to the
  invariance properties of  recognition systems. Invariance can be
  engineered (e.g.~convolutional nets), or learned from data
  explicitly (e.g.~temporal coherence) or implicitly (e.g.~by data
  augmentation). One idea that has not, to date, been explored is the
  integration of latent variables which permit a search over a learned
  space of transformations. Motivated by evidence that people mentally
  simulate transformations in space while comparing examples,
  so-called ``mental rotation'', we propose a \emph{transforming
    distance}. Here, a trained relational model actively
  transforms pairs of examples so that they are maximally similar
  in some feature space yet respect the learned transformational
  constraints. We apply our method to nearest-neighbour problems on
  the Toronto Face Database and NORB.
\end{abstract}

\section{Introduction}

It has long been conjectured that humans, when tasked with recognizing
the identity of an object or judging the similarity between objects,
mentally simulate transformations of internal representations of those
objects. Shepard and Metzler \cite{shepard1971mental} were the first
to formalize this phenomenon, and assess it experimentally. They
presented people with two line drawings, each of which portrayed a
three-dimensional object in space.  They showed that the reaction time
for participants to decide whether the objects were the same (except
for a rotation in space) or were different, was linearly related to
the angle between the spatial orientation of the two objects. This
finding indicated that a type of ``mental rotation'' was performed. If
instead, a rotationally invariant representation of each object was
formed, then the time required for a decision would presumably be
independent of angle.

Although much progress on object recognition by machines has been
inspired by human biology \cite{pinto2008real}, these models have rarely
accounted for the explicit transformation of internal representations
analogous to human mental rotation. Instead, much focus has been
placed on developing recognizers that are invariant to spatial
transformations. One example where invariance is engineered into the
model is convolutional networks \cite{LeCun1998}, which gain
invariance to small translations in the input because they pool
features over local regions. An alternative is to \emph{learn}
invariance, for example by augmenting the training set with
perturbations of the training set
\cite{bengio2011deep,Ciresan:2011:FHP:2283516.2283603,paulin:hal-00979464},
through temporal cues \cite{foldiak1991learning,
  mobahi2009deep,liao2013learning} or incorporating linear
transformation operators into feature learning algorithms
\cite{sohn2012learning}. Additionally, canonical correlation analysis (CCA)
\cite{hotelling1936relations} and its non-linear variants have been used
to model relationships between example pairs, including images of the 
same object under different view angles \cite{melzer2003appearance}
and images under different illumination conditions \cite{gutmann2014spatio}.
Another related stream includes various deformable part models 
\cite{felzenszwalb2010object} and diffeomorphism models 
\cite{sparks2010novel}, where knowledge about specific spatial relationships or classes of transformations are encoded.

Some non-parametric methods bear a resemblance to  mental 
rotation, for example, nearest-neighbor techniques in which examples
are explicitly compared. However, for such methods to work well in
domains with large intra-class variance, one either needs to maintain
a database of essentially all different appearances of objects, or learn an
invariant similarity measure \cite{Goldberger2004,Hadsell2006}.

A class of relational unsupervised learning techniques
\cite{memisevic2013learning} use multiplicative interactions between
inputs to represent correlation patterns across
multiple images. One application of these methods has been to learn to
represent transformations between image pairs
\cite{memisevic2007unsupervised}. Such a model can be used to assess
the similarity between images and used for nearest-neighbor
classification \cite{susskind2011modeling}. 
Another related line of research focuses on disentangling different 
attributes \cite{desjardins2012disentangling, reed2014learning}, 
for example face identity and expression. These models have multiple 
groups of hidden units corresponding to one group of visible units, 
where each group learns `absolute' representation of one attribute.
In contrast, relational models 
\cite{memisevic2013learning,memisevic2007unsupervised}
learn to encode `relative' transformations between multiple groups 
of visible units, where each group represents one transformed instance
of an image.

In this paper, we propose a neural architecture, inspired by mental
rotation, that explicitly transforms representations of images while
evaluating their similarity. The two key components of our approach
are i) a relational model trained on pairs of images which learns
about the range of valid transformations; and ii) an optimization
that attempts to transform an image such that its representation
matches that of a target while respecting the constraints learned by
the relational model.
%
\section{Background}
Our work attempts to link two areas of study which have to-date
remained disparate: unsupervised learning of image transformations and
learning a similarity measure such that objects that are perceptually
similar will have a high measurable similarity.

The latter field has focused on methods which are computationally
tractable, often learning a Mahalanobis distance
\cite{davis2007information} or other functional mapping
\cite{Goldberger2004, Hadsell2006} in which distances can be computed
for tasks such as nearest neighbour classification or document
retrieval. Such methods typically exploit distance information that is
intrinsically available: binary or real-valued similarity or
dissimilarity labels, user input, or class-based information. Rarely
has the set of known transformations of an object been incorporated in
learning. An exception is Hadsell et al.~\cite{Hadsell2006}, though
they do not attempt to model transformations. However, transformations
are a cue for learning an invariant measure and often can be acquired
cheaply through unlabeled data, for example, video.


Characterizing a transformation is central to our task. The field of
representation learning \cite{bengioreview} is concerned with learning
features which untangle unknown underlying factors of variation in
data. These representations enable higher-level reasoning without
domain-specific engineering. While the majority of these techniques
are concerned with learning representations of individual objects
(e.g.~images) a subclass of these methods aim to learn relations from
pairs of objects \cite{memisevic2007unsupervised,
  memisevic2010learning, karklin2009emergence}.  These techniques have
been applied to feature covariances in image and audio data
\cite{ranzato2010modeling,dahl2010phone}, learning image
transformations \cite{memisevic2007unsupervised, susskind2011modeling,
  memisevic2011gradient}, and spatio-temporal features for activity
recognition \cite{taylor2010convolutional,konda2013role}.

Both probabilistic \cite{memisevic2007unsupervised,
  memisevic2010learning}) and non-probabilistic
\cite{memisevic2011gradient} variants of relational feature learning
methods exist. The former is based on extending the restricted
Boltzmann machine to three-way rather than pairwise interactions. The
latter extends auto-encoders in a similar way. Although the
auto-encoder-based approach is easier to train using gradient descent,
we adopt the RBM formalism in our work due to the simplicity with
which its free energy can be computed to score input pairs under the
model. Recent work also suggests a means of scoring inputs under an
autoencoder \cite{kamyshanska2013autoencoder}.
\section{Methods}
%
%
A relational model captures the transformation between meaningful
pairs of images $(\bm{x}, \bm{y})$, such as photos taken from
different views of the same object, and images with different
expressions of the same face. It encodes the transformation
information of $\bm{x}$ to $\bm{y}$ as a hidden representation $\bm{h}$,
which can also be used to transform the ``source'' $\bm{x}$ towards the ``target'' $\bm{y}$.

\subsection{Factored gated restricted Boltzmann machine}

\def\imgwidth{1.3cm}
\def\subwidth{0.3\linewidth}
\begin{figure}[t]
\vspace{-.75cm} 
\begin{minipage}[b]{\subwidth}
\begin{tikzpicture}
\begin{scope}[thick, node distance=2.5em, on grid
    ]
\node[block](h){$\bm{h}$};
\node[factor](fac) [below=of h] {} edge [-] (h);
\node[block](x)[below=of fac,xshift=-4em]{$\bm{x}$} edge [->] (fac.left corner);
\node[block](y)[right=of x, xshift=5em] {$\bm{y}$} edge [-] (fac.right corner);
\end{scope}
\end{tikzpicture}
\subcaption{Factored gated RBM}
\end{minipage}
\begin{minipage}[b]{\subwidth}
\centering
\begin{minipage}[b]{\imgwidth}
\includegraphics[width=\imgwidth]{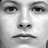}
\end{minipage}
\begin{minipage}[b]{\imgwidth}
\includegraphics[width=\imgwidth]{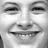}
\end{minipage}

\begin{minipage}[b]{\imgwidth}
\includegraphics[width=\imgwidth]{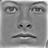}
\end{minipage}
\begin{minipage}[b]{\imgwidth}
\includegraphics[width=\imgwidth]{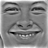}
\end{minipage}

\subcaption{Facial expression pair}
\label{sameidentity}
\end{minipage}
\begin{minipage}[b]{\subwidth}
\centering
\begin{minipage}[b]{\imgwidth}
\includegraphics[width=\imgwidth]{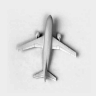}
\end{minipage}
\begin{minipage}[b]{\imgwidth}
\includegraphics[width=\imgwidth]{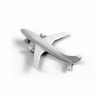}
\end{minipage}

\begin{minipage}[b]{\imgwidth}
\includegraphics[width=\imgwidth]{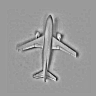}
\end{minipage}
\begin{minipage}[b]{\imgwidth}
\includegraphics[width=\imgwidth]{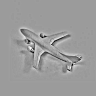}
\end{minipage}

\subcaption{\centering{Same object pair}}
\label{sameobject}
\end{minipage}

\caption{Relational architecture used for learning transformation as
  well as examples of image pairs used to train the model. In subfigures (b) and (c), the top rows contain original pairs of images and the bottom rows contain processed pairs of images after local contrast normalization.}
\label{fig: fgrbm}
\end{figure}
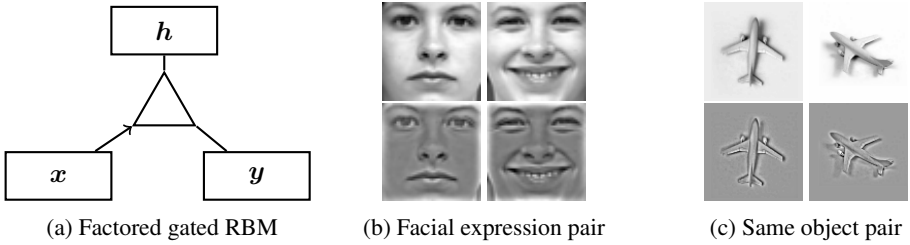
%
The factored gated RBM (fgRBM) \cite{memisevic2010learning} relates
$\bm{x}$, $\bm{y}$ and $\bm{h}$ by the following energy function:
\begin{equation}
\label{eq: gaussian_fgrbm_energy}
-E(\bm{y}, \bm{h}; \bm{x}) = \sum_{n=1}^N (\sum_{i=1}^I v_{in} x_i) (\sum_{j=1}^J w_{jn} y_j) (\sum_{m=1}^M u_{mn} h_m) + \frac{1}{2}\sum_{j=1}^J (y_j - b_j)^2 + \sum_{l=m}^M c_m h_m ~,
\end{equation}
where $I$, $J$ and $M$ are the dimensionality of $\bm{x}$, $\bm{y}$
and $\bm{h}$ respectively; $v_{in}$, $w_{jn}$, and $u_{ln}$ are a set
of filters that project $\bm{x}$, $\bm{y}$ and $\bm{h}$ onto $N$
factors. 
$b_j$ and
$c_m$ are the bias coefficients for $\bm{y}$ and $\bm{h}$ respectively.
Equation \ref{eq: gaussian_fgrbm_energy} defines a Gaussian-Bernoulli
\cite{hinton2006reducing} version of fgRBM capable of modeling
real-valued data, which has a slightly different energy function from
that of the original binary fgRBM in
\cite{memisevic2010learning}. Equation \ref{eq: gaussian_fgrbm_energy}
assumes that $\bm{x}$ and $\bm{y}$ are real-valued and $\bm{h}$ is
binary.  

We express the probability of $\bm{y}$ and $\bm{h}$, conditioned on $\bm{x}$, in terms of the defined energy function as:
\begin{equation}
\label{eq: joint_prob}
p(\bm{y}, \bm{h}; \bm{x}) = \frac{1}{Z(\bm{x})} \mathrm{exp}(-E(\bm{y}, \bm{h}; \bm{x})),\
\end{equation}
where the partition function $Z(\bm{x}) = \sum_{\bm{y}, \bm{h}}
\mathrm{exp}(-E(\bm{y}, \bm{h}; \bm{x}))$ ensures normalization.
Marginalizing over $\bm{h}$, we obtain the probability distribution of $\bm{y}$:
\begin{align}
\label{eq: y_prob}
p(\bm{y}; \bm{x}) &= \sum_{\bm{h} \in \{0, 1\}^M} p(\bm{y}, \bm{h}; \bm{x})  = \frac{1}{Z(\bm{x})} \mathrm{exp}(-F(\bm{y}; \bm{x})),\
\end{align}
where $F(\bm{y}; \bm{x})$ is the free energy of $\bm{y}$, which is
defined as $-\mathrm{log} \sum_{\bm{h} \in \{0, 1\}^M}
\mathrm{exp}(-E(\bm{y}, \bm{h}; \bm{x}))$, or
\begin{align}
\label{eq: free_energy}
F(\bm{y}; \bm{x}) &= -\frac{1}{2}\sum_j (y_j - b_j)^2 - \sum_l \mathrm{log}(1 + \mathrm{exp}(\sum_{f} u_{mn}\sum_{i} v_{in} x_i \sum_{j} w_{jn} y_j   + c_m)).\
\end{align}


%

\subsubsection{Factored gated RBM training}

To train the fgRBM, we would like to maximize the average log
probability of $\bm{y}$ given $\bm{x}$ for a set of training pairs
$\{(\bm{x}^{\alpha}, \bm{y}^{\alpha})\}$:
\begin{equation}
\mathcal{L}(\bm{\theta}) = \sum_{\alpha} \mathrm{log} p(\bm{y}^{\alpha}; \bm{x}^{\alpha})
\end{equation}
where $\bm{\theta}$ represents all the involved parameters.

The partial derivative of the likelihood  $\mathcal{L}(\bm{\theta})$ with respect
to any parameter $\bm{\theta}$ is
%
\begin{align}
\label{eq: CD_theta_E}
-\frac{\partial \mathcal{L}}{\partial \bm{\theta}} & = 
\sum_{\alpha} \langle \frac{\partial E(\bm{y}^{\alpha}, \bm{h}; \bm{x}^{\alpha})}{\partial \bm{\theta}} \rangle_{\bm{h}} - \langle \frac{\partial E(\bm{y}, \bm{h}; \bm{x}^{\alpha})}{\partial \bm{\theta}} \rangle_{\bm{y, h}}
= \sum_{\alpha} \frac{\partial F(\bm{y}^{\alpha}; \bm{x}^{\alpha})}{\partial \bm{\theta}} - \langle \frac{\partial F(\bm{y}; \bm{x}^{\alpha})}{\partial \bm{\theta}} \rangle_{\bm{y}}
\end{align}

where $\langle \rangle_z$ denotes the expectation with respect to $z$.
The first term is the derivative of the free energy which is tractable
to compute. The second term involves integrating over all possible
$\bm{y}$ and therefore is intractable in non-trivial settings. In this
paper, we used the contrastive divergence learning algorithm
\cite{hinton2002training}. 

\subsection{Transforming distance}

We can use a trained fgRBM to transform an image
$\bm{x}$ with a given $\bm{h}$ which encodes a transformation. The
transforming function can be derived from the conditional $p(\bm{y}|\bm{h};\bm{x})$
by taking the conditional expectation of $\bm{y}|\bm{x}$. We use
$\bm{t}(\bm{x}, \bm{h})$ to represent the transformed $\bm{x}$ with
respect to $\bm{h}$. $\bm{t}(\bm{x}, \bm{h})$ has the same
dimensionality as $\bm{y}$ and we express the j\textsuperscript{th}
element of $\bm{t}$ as
\begin{equation}
\label{eq: trans_image}
t_j(\bm{x}, \bm{h}) = \sum_{n} w_{jn}\sum_{i} v_{in} x_i \sum_{m} u_{mn} h_m + b_j ~.
\end{equation}

With  $\bm{t}(\bm{x}, \bm{h})$, we can define the transforming distance $D$ between a pair of images $(\bm{x}^{\alpha}, \bm{x}^{\beta})$, given their transformations represented by $\bm{h}^{\alpha}$ and $\bm{h}^{\beta}$ respectively:
\begin{align}
D_s(\bm{x}^{\alpha}, \bm{x}^{\beta}, \bm{h}^{\alpha}) &= d\left(f\left(\bm{t}\left(\bm{x}^{\alpha}, \bm{h}^{\alpha}\right)\right), f\left(\bm{x}^{\beta}\right)\right) ~,\\
D_d(\bm{x}^{\alpha}, \bm{x}^{\beta}, \bm{h}^{\alpha}, \bm{h}^{\beta}) &= d\left(f\left(\bm{t}\left(\bm{x}^{\alpha}, \bm{h}^{\alpha}\right)\right), f\left(\bm{t}\left(\bm{x}^{\beta}, \bm{h}^{\beta}\right)\right)\right) ~,
\end{align}
where $D_s$ represents the transforming distance with a
\emph{single-sided} transformation and $D_d$ uses a \emph{dual-sided}
transformation. Intuitively, the difference between these approaches
is whether $\bm{x}^{\alpha}$ is transformed to match $\bm{x}^{\beta}$
or if $\bm{x}^{\alpha}$ and $\bm{x}^{\beta}$ are both transformed to
match each other.

\begin{figure}[t]
\vspace{-.75cm}
\begin{minipage}[b]{0.42\linewidth}
\begin{tikzpicture}
\begin{scope}[thick, node distance=2.5em, on grid
    ]
\node[block](h){$\bm{h}^\alpha$};
\node[factor](fac) [below=of h] {} edge [-] (h);
\node[block](x_0)[below=of fac,xshift=-3em]{$\bm{x}^\alpha$} edge [->] (fac.left corner);
\node[block](y)[right=of x_0, xshift=4em] {$\bm{t}(\bm{x}^{\alpha}, \bm{h}^{\alpha})$} edge [-] (fac.right corner);

\node[sblock](ftr_0)[above=of y, xshift=0.8em] {$f(\bm{t}^\alpha)$} edge [-] (y);

\node[block](dist)[above=of ftr_0, xshift=1.7em, yshift=0.5em]{$D(\bm{x}^{\alpha}, \bm{x}^{\beta}, \bm{h})$} edge[-](ftr_0);

\node[sblock](ftr_1)[below=of dist, xshift=1.7em, yshift=-0.5em] {$f(\bm{x}^\beta)$} edge [-] (dist);

\node[block](x_1)[right=of y, xshift=2.5em]{$\bm{x}^\beta$} edge[-](ftr_1);
\end{scope}
\end{tikzpicture}
\subcaption{Single-sided transformation}
\label{singletrans}
\end{minipage}
\begin{minipage}[b]{0.58\linewidth}
\begin{tikzpicture}
\begin{scope}[thick, node distance=2.5em, on grid
    ]
\node[block](h_a){$\bm{h}^\alpha$};
\node[factor](fac_a) [below=of h_a] {} edge [-] (h_a);
\node[block](x_0)[below=of fac,xshift=-3em]{$\bm{x}^\alpha$} edge [->] (fac_a.left corner);
\node[block](t_a)[right=of x_0, xshift=4em] {$\bm{t}(\bm{x}^{\alpha}, \bm{h}^{\alpha})$} edge [-] (fac_a.right corner);

\node[sblock](ftr_a)[above=of t_a, xshift=0.8em] {$f(\bm{t}^\alpha)$} edge [-] (t_a);
\node[block](dist)[above=of ftr_a, xshift=1.8em, yshift=0.5em]{$D(\bm{x}^{\alpha}, \bm{x}^{\beta}, \bm{h})$} edge[-](ftr_a);

\node[sblock](ftr_b)[right=of ftr_a, xshift=0.8em] {$f(\bm{t}^\beta)$} edge [-] (dist);

\node[block](h_b)[right=of h_a, xshift=9.75em]{$\bm{h}^\beta$};
\node[factor](fac_b) [right=of fac_a, xshift=9.25em] {} edge [-] (h_b);
\node[block](t_b)[right=of t_a, xshift=2.5em]{$\bm{t}(\bm{x}^{\beta}, \bm{h}^{\beta})$} edge[-](ftr_b) edge[-](fac_b.left corner) ;
\node[block](x_b_0)[below=of fac_b,xshift=3em]{$\bm{x}^\beta$} edge [->] (fac_b.right corner);
\end{scope}
\end{tikzpicture}
\subcaption{Dual-sided transformation}
\label{dualtrans}
\end{minipage}

\caption{Transforming distance. (\subref{singletrans}) in a
  single-sided transformation $\bm{x}^{\alpha}$ is
transformed to match $\bm{x}^{\beta}$. (\subref{dualtrans}) in a
dual-sided transformation $\bm{x}^{\alpha}$
and $\bm{x}^{\beta}$ are both transformed to match each other.}
\label{fig: trans_dist}
\end{figure}
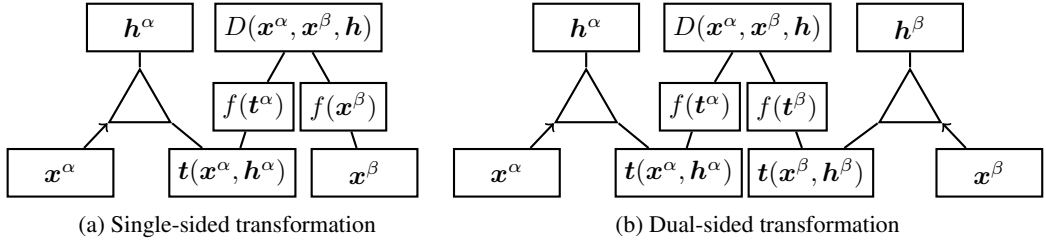

$d(\bm{p}, \bm{q})$ can be any pairwise distance and $f(\bm{q})$ can
be any feature representation of $\bm{q}$. In this paper, for distance
metrics, we simply consider the Euclidean distance; for feature
representations, we consider raw pixels, principal component analysis
(PCA) and the contractive autoencoder (CAE)
\cite{rifai2011contractive}.  Diagrams for both single and dual-sided
transformed distances are shown in Fig.~\ref{fig: trans_dist}.

By doing ``mental rotation'' during image comparison, images undergo
learned transformations to minimize their distance.  If we only
minimize the transforming distance $D$ over $\bm{h}$, the transformed
image $\bm{t}(\bm{x}^{\alpha}, \bm{h}^{\alpha})$ might be very similar
to image $\bm{x}^{\beta}$ while not respecting the constraints learned
by the fgRBM, i.e. not being a ``valid'' transformation of
$\bm{x}^{\alpha}$. To effectively tie the ``identity'' of
$\bm{t}(\bm{x}^{\alpha}, \bm{h}^{\alpha})$ to $\bm{x}^{\alpha}$
throughout the transformation, we regularize this optimization by its
free energy defined in Equation \ref{eq: free_energy}. Accordingly, we
have the following cost function:
\begin{equation}
\label{eq: cost_opt}
L(\bm{x}^{\alpha}, \bm{x}^{\beta}, \bm{h}) = D(\bm{x}^{\alpha}, \bm{x}^{\beta}, \bm{h}) + \lambda R(\bm{x}^{\alpha}, \bm{x}^{\beta}, \bm{h}) ~,    
\end{equation}
where $\bm{h}$ represents both $\bm{h}^\alpha$ and $\bm{h}^\beta$ for
brevity. Regularization term $R$ represents the free energy term(s):
for a single-sided transformation, $R(\bm{x}^{\alpha}, \bm{h}) =
F(\bm{t}(\bm{x}^{\alpha}, \bm{h}); \bm{x}^{\alpha})$; for a dual-sided
transformation $R(\bm{x}^{\alpha}, \bm{h}) = F(\bm{t}(\bm{x}^{\alpha},
\bm{h}); \bm{x}^{\alpha}) + F(\bm{t}(\bm{x}^{\beta}, \bm{h});
\bm{x}^{\beta})$. $\lambda$ is an empirically chosen weight constant
to balance $D$ and $R$ (we set $\lambda = 1$ in all of our experiments).

The optimized fgRBM hidden vector $\bm{h}^{*}$ is defined as:
\begin{equation}
\label{eq: h_opt}
\bm{h}^{*}(\bm{x}^{\alpha}, \bm{x}^{\beta}) = \argmin_{\bm{h}} L(\bm{x}^{\alpha}, \bm{x}^{\beta}, \bm{h}) ~.
\end{equation}
Instead of optimizing the binary hidden units directly, we optimize
the real-valued logits (pre-sigmoid inputs). 
The pairwise image distance of $(\bm{x}^{\alpha}, \bm{x}^{\beta})$ after mental rotation is defined as:
\begin{equation}
D^{*}(\bm{x}^{\alpha}, \bm{x}^{\beta}) = D(\bm{x}^{\alpha},
\bm{x}^{\beta}, \bm{h}^{*}(\bm{x}^{\alpha}, \bm{x}^{\beta})) ~.
\end{equation}
%
%
We use gradient descent with momentum to compute Equation \ref{eq:
  h_opt}.  
To allow parallel processing over pairs on a GPU, we do not set a
stopping criterion for individual pairs. We stop the optimization
after 30 iterations. We note that more sophisticated gradient-based
optimization methods can be used, though they may not be as amenable
to GPU-based parallelization.

\section{Experiments}

We considered two datasets: the Toronto Face
Database (TFD) \cite{susskind2010toronto} and the NORB dataset
\cite{lecun2004learning}. A qualitative visual assessment was
performed to demonstrate the validity of optimizing the transforming
distance. To quantitatively evaluate, we replaced the L2 distance
in K nearest neighbor (KNN) algorithm with the proposed transforming 
distance. We denote this variant of KNN as \emph{transforming KNN}. 
Specifically, a fgRBM is trained at the training stage; during test time, 
this fgRBM defines a transforming distance that is used to match 
test images with examples in the KNN database.

\subsection{Data}
\label{sec: exp_data}

{\bf The TFD} contains 102,236 face images of size 48 $\times$ 48.  
3,874 of them are labeled with identities of 963 people.  Among these,
626 identities (3,537 images) have at least 2 images for each
identity. We refer to this subset as the identity-labeled images.
We first learned the valid expression transformation within the same
face identity using the fgRBM (see Fig.~\ref{sameidentity}). We expect
the transforming distance to be expression invariant and evaluate this
by performing identity recognition.

The TFD dataset was divided into four sets:
\begin{itemize}
    \item the feature training set, which contains all the
      identity-unlabeled images and identities with a single image
      (separated so that features are not biased towards either training or test set);
    \item the KNN training set (database), used for both training the
      fgRBM and as a KNN database;
    \item the KNN validation set, used for hyper-parameter cross-validation;
    \item and the KNN test set.
\end{itemize}

To the best of our knowledge, we are unaware of any reported work on identity recognition using the TFD dataset. Therefore, we explore two methods to divide the identity-labeled images into training, validation and test sets, refered as TFDs1 and TFDs2:
\begin{itemize}
    \item \textbf{TFDs1}: For each identity, randomly take 1 image as test. 
    For the remaining data, 
    randomly take 1 as validation from any identity with at least 2 images.
    Each random division results in a training set of size 2,540, 
    a validation set of size 371 and a test set of size 626.
    The training set contains 21,948 image pairs, including self pairs
    (i.e.~the identity transformation).
  \item \textbf{TFDs2}: For each identity with at least 4 images,
    randomly take 1 image as test and 1 as validation.  Each division
    results in a training set of size 2,937, a validation set of size
    300 and a test set of size 300.  The training set contains 23,607
    image pairs, including self pairs.
\end{itemize}


There are several differences between TFDs1 and TFDs2.  In TFDs1, the
test set is larger.  However, 326 identities have only one
example.  This challenges the model in two ways: i) identities cannot
be learned by the fgRBM; ii) each identity appears only once
 in the KNN database.  In the TFDs2 training set, images always
appear in pairs, which favors both fgRBM training and KNN.  However,
since only identities with at least 4 instances contribute to the
validation and test sets, the KNN database is ``diluted'' by  891 ``extra'' images which
share identity with neither validation nor test.

{\bf The NORB dataset} contains $96 \times 96$ stereo image pairs
of 50 toys from 5 categories. Images were taken under different
lighting conditions, elevations and azimuths
\cite{lecun2004learning}. We used the ``small NORB''
variant, where the images have a clean background. The training and
test sets each contain 24,300 images.  For NORB, we try to model the
transformation between images of the same object with different
azimuth (see Fig.~\ref{sameobject}). 
We expect the transforming distance to be
rotation invariant.\looseness=-1

We arbitrarily assigned all images with instance number 7 to the
validation set, which includes 4,860 images, and all images with
instance label 4, 6, 8, 9 to the training set, which contains 19,440
images.  The training set was used for feature learning, fgRBM training
and as the KNN database. We used the default test set split.

\subsection{Model training}

Before training the fgRBM and performing transforming distance, TFD
images were preprocessed by local contrast normalization (LCN)
\cite{pinto2008real} with kernel size 9. For the NORB dataset, we only
used the first image from each stereo pair. The images were first
down-sampled to $32 \times 32$ and then preprocessed by LCN with
kernel size 3. Examples are shown in Fig.~\ref{fig: fgrbm} (b) and
(c).

For TFD, the fgRBM is trained on image pairs with the same identity.
The fgRBM encodes valid expression transformation information.  For
NORB, within each image pair, 2 images share the same class, instance,
lighting condition and elevation.  The fgRBM encodes the valid azimuth
transformation information.  Based on the allowed azimuth difference
between images, we trained 2 types of models: 1) \emph{AnyTrans},
where the images within a pair have an arbitrary azimuth difference;
and 2) \emph{SmallTrans}, where the absolute value of azimuth
difference is less than or equal to 40$^\circ$.

\subsubsection{Hyperparameter selection}

For the feature representation, we cross-validated the choice of PCA
and CAE with 64, 128 and 256 hidden units on regular KNN.  The best
one of these 6 combinations was chosen.  For the fgRBM, we
cross-validated hidden dimensionality of 64, 128 and 256 by performing
a single-sided transforming KNN with pixel distance. For NORB, the $K$
value for KNN was cross-validated in the range of 1 to 30.  While for
TFD, $K$ was set to 1 because of the nature of the task. The number of
factors in the fgRBM is fixed as twice the dimensionality of the
hidden units. Due to space constraints, details of the remaining
settings are provided in supplemental material.



\subsection{Qualitative evaluation} 

\begin{figure}[t]
\vspace{-.75cm} 
\centering
\includegraphics[width=\imgwidth]{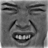}
\includegraphics[width=\imgwidth]{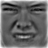}
\includegraphics[width=\imgwidth]{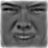}
\includegraphics[width=\imgwidth]{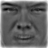}
\includegraphics[width=\imgwidth]{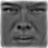}
\includegraphics[width=\imgwidth]{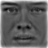}
\includegraphics[width=\imgwidth]{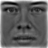}
\includegraphics[width=\imgwidth]{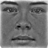}

\includegraphics[width=\imgwidth]{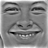}
\includegraphics[width=\imgwidth]{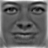}
\includegraphics[width=\imgwidth]{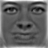}
\includegraphics[width=\imgwidth]{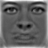}
\includegraphics[width=\imgwidth]{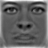}
\includegraphics[width=\imgwidth]{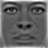}
\includegraphics[width=\imgwidth]{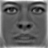}
\includegraphics[width=\imgwidth]{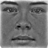}

\includegraphics[width=\imgwidth]{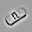}
\includegraphics[width=\imgwidth]{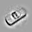}
\includegraphics[width=\imgwidth]{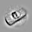}
\includegraphics[width=\imgwidth]{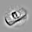}
\includegraphics[width=\imgwidth]{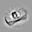}
\includegraphics[width=\imgwidth]{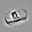}
\includegraphics[width=\imgwidth]{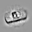}
\includegraphics[width=\imgwidth]{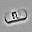}

\includegraphics[width=\imgwidth]{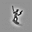}
\includegraphics[width=\imgwidth]{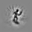}
\includegraphics[width=\imgwidth]{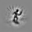}
\includegraphics[width=\imgwidth]{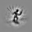}
\includegraphics[width=\imgwidth]{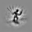}
\includegraphics[width=\imgwidth]{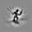}
\includegraphics[width=\imgwidth]{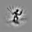}
\includegraphics[width=\imgwidth]{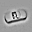}
\caption{Image sequences illustrating the optimization process of the single-sided transforming distance. 
In each row, the first image is the ``source'' image $\bm{x}^\alpha$; 
the last image is the ``target'' image $\bm{x}^\beta$; 
images in-between are transformations of the source image $\bm{t}(\bm{x}^{\alpha}, \bm{h}^{\alpha})$, 
the second last image is the optimized transformation $\bm{t}(\bm{x}^{\alpha}, \bm{h}^{*})$.
The 1st and 3rd rows show the process of transforming same-identity pairs.
The 2nd and 4th rows show different-identity
pairs. Note preservation of identity.\looseness=-1
}
\label{fig: img_seq}
\end{figure}

To assess the transforming distance optimization process, we observed
the transformed image $\bm{t}(\bm{x}^{\alpha}, \bm{h}^{*})$ at each
optimization step. Fig.~\ref{fig: img_seq}
shows the optimization process of both same-identity 
and different-identity 
image pairs.  We see that during the optimization process, the
transformed images are all valid transformations of the source images:
the faces images preserves their identities, even when an image is
transformed to match a face image with different identity; the car and
human figure are merely rotated, even when the human figure image is
transformed to match the car image.  This is an interesting
generalization result, since the fgRBM was only trained on
same-identity pairs.

\subsection{Quantitative evaluation}

For quantitative evaluation, the transforming distance was combined with KNN 
as the transforming KNN. 
It is compared with 2 baseline methods: 
the regular KNN and a proposed \emph{augmented KNN}.

\subsubsection{Augmented KNN}

By performing ``space searching'' when calculating transforming distances, 
each example in KNN database covers a volume, 
a manifold composed of all its valid transformations, in the data space.
It is interesting to see if this ``expanding effect'' 
can be achieved by augmenting the KNN database 
using the learned transformation and then performing regular KNN.
This comparison is meaningful especially when the data space is not
well covered by the KNN database (e.g.~in the TFD dataset). 

Therefore, we proposed the ``augmented KNN'' and test it on
TFD. Additional training images were sampled from a trained fgRBM.
We used each image in the KNN database as the ``source''
image, and performed Gibbs sampling alternating between hidden
and target units.  We retained one sample every 100 iterations.  By
repeating this 9 times for every image in the KNN database, we
acquired a database 10 times as large as the original database.  We
are unaware of any prior work which has used a relational model for
dataset augmentation.  Provided the data to train the relational model
is available, this seems like an attractive option to the usual
approach of hand-coding perturbations \cite{bengio2011deep,
  Ciresan:2011:FHP:2283516.2283603}.\looseness=-1

\subsubsection{KNN performance}

\begin{table}[t]
\vspace{-.75cm} 
\small
\centering
\begin{tabular}{cccccccc}
\hline
 Data & \multicolumn{2}{c}{Regular KNN} & \multicolumn{2}{c}{Augmented KNN} & \multicolumn{2}{c}{Transforming KNN} &  Type of\\
\cline{2-7}
 separation  & pixel & feature & pixel & feature & pixel & feature & transformation\\
\hline

\multirow{2}{*}{TFDs1} & \multirow{2}{*}{29.7$\pm$1.5}  & \multirow{2}{*}{31.2$\pm$1.2} & \multirow{2}{*}{32.2$\pm$0.9} & \multirow{2}{*}{33.3$\pm$0.6} &  36.9$\pm$1.3  & 37.7$\pm$1.4 & single\\
& & & & & 38.6$\pm$0.9 & \textbf{39.2}$\pm$0.3 & dual\\

\multirow{2}{*}{TFDs2} & \multirow{2}{*}{41.1$\pm$1.3}  & \multirow{2}{*}{44.6$\pm$1.6}  & \multirow{2}{*}{47.4$\pm$1.7}& \multirow{2}{*}{49.5$\pm$2.1}& 67.0$\pm$3.2  & 68.9$\pm$2.9 & single\\

    &  & & &   & \textbf{69.5}$\pm$2.7  & \textbf{69.8}$\pm$2.8   & dual\\

\hline
\end{tabular}
\caption{KNN classification accuracy (\%) on the TFD dataset. 
}

\label{tab: tfd}
\end{table}

\begin{table}[t]
\small
\centering
\begin{tabular}{cccccc}
\hline
Value & \multicolumn{2}{c}{Regular KNN} & \multicolumn{2}{c}{Transforming KNN} &  Type of\\
\cline{2-5}
of $K$ & pixel & feature & pixel & feature & rotation\\
\hline
\multirow{2}{*}{cross-validated} &\multirow{2}{*}{78.4} &\multirow{2}{*}{78.8} &80.7 & 71.4 & any\\
 & & & \textbf{83.0} & 76.7&   small\\


\multirow{2}{*}{1} &\multirow{2}{*}{78.4} &\multirow{2}{*}{78.8} &76.2 & 67.1 & any\\
& & & \textbf{80.4} & 74.5 &   small\\



\hline
\end{tabular}
\caption{KNN classification accuracy (\%) on the NORB dataset. 
}
\label{tab: norb}
\vspace{-.5cm} 
\end{table}

\textbf{TFD} results are shown in Table \ref{tab: tfd}.
For both TFDs1 and TFDs2, we analyzed both single and dual-sided
transforming KNN. Results are compared against augmented KNN and regular
KNN. Transforming KNN brings 6 -- 10 \% performance
increase for TFDs1 and about 25 \% performance increase for TFDs2.
Augmented KNN also brings some amount of accuracy increase, 
but not as high as transforming KNN.
This indicates that the augmented database does not cover as much as
volume as the transforming distance, at least when the database is
expanded tenfold.

The performance differences between TFDs1 and TFDs2 correspond
 to our analysis in Section \ref{sec: exp_data}. 
Accuracies in every column is higher in TFDs2 than in TFDs1. 
This is probably due to the fact that in TFDs2, 
every test image has at least two corresponding images in the KNN database, 
while in TFDs1, about half of the test images only have one. 
The relative performance increase is higher in TFDs2 than in
TFDs1. This is probably because that in TFDs2, expression
transformation was learned for every identity, while in TFDs1, this
was done only on about half of the identities.

\textbf{NORB} results are given in Table \ref{tab: norb}.  With a
cross-validated $K$ value, the SmallTrans KNN has a 4\% performance
increase over regular KNN. We are not surprised by the marginal
improvement of the transforming KNN.  This is because the NORB
training set contains images of the same object taken from different
azimuths (every 20$^\circ$), which already provides some degree of
rotation invariance.  Therefore SmallTrans and AnyTrans can only do
slightly better, if they provide a finer rotation invariance, say
rotations under 20$^\circ$ difference, for some cases. However, the
rotation invariance afforded by the training set disappears when there
are less examples in the database, which is further illustrated in
Fig.~\ref{fig: knn_reduce} and discussed in Section \ref{sec:
  exp_knn_reduce}.

Fig.~\ref{fig: knn_k} shows KNN accuracy with respect to different $K$
values.  For TFD, we can see that both regular KNN and transforming
KNN degrade as $K$ increases.  This is mainly due to the lack of
examples in its KNN database: if $K$ is large, number of noise
examples will be larger than true examples even the true examples
might have higher similarities, which is also aggravated by the
identity-imbalanced nature of TFD.  In comparison, augmented KNN is
the most robust to larger settings of $K$, due to the richness of same-identity
examples in the KNN database.

For NORB, regular KNN accuracy decreases with increasing $K$.  This is
because without transforming distance, a test image should only be
close to examples with the same class label and, with equal
importance, the orientation in the KNN database.  Therefore, test images
cannot utilize many examples to provide effective distances.  On
the contrary, transforming KNN can utilize training examples
regardless of their orientation, which results in an increasing
robustness with increasing $K$.

\def\figwidth{4.75cm}
\begin{figure}[t]
\vspace{-.75cm} 
\centering
\begin{minipage}[c]{0.32\linewidth}
\centering
\includegraphics[width=\figwidth]{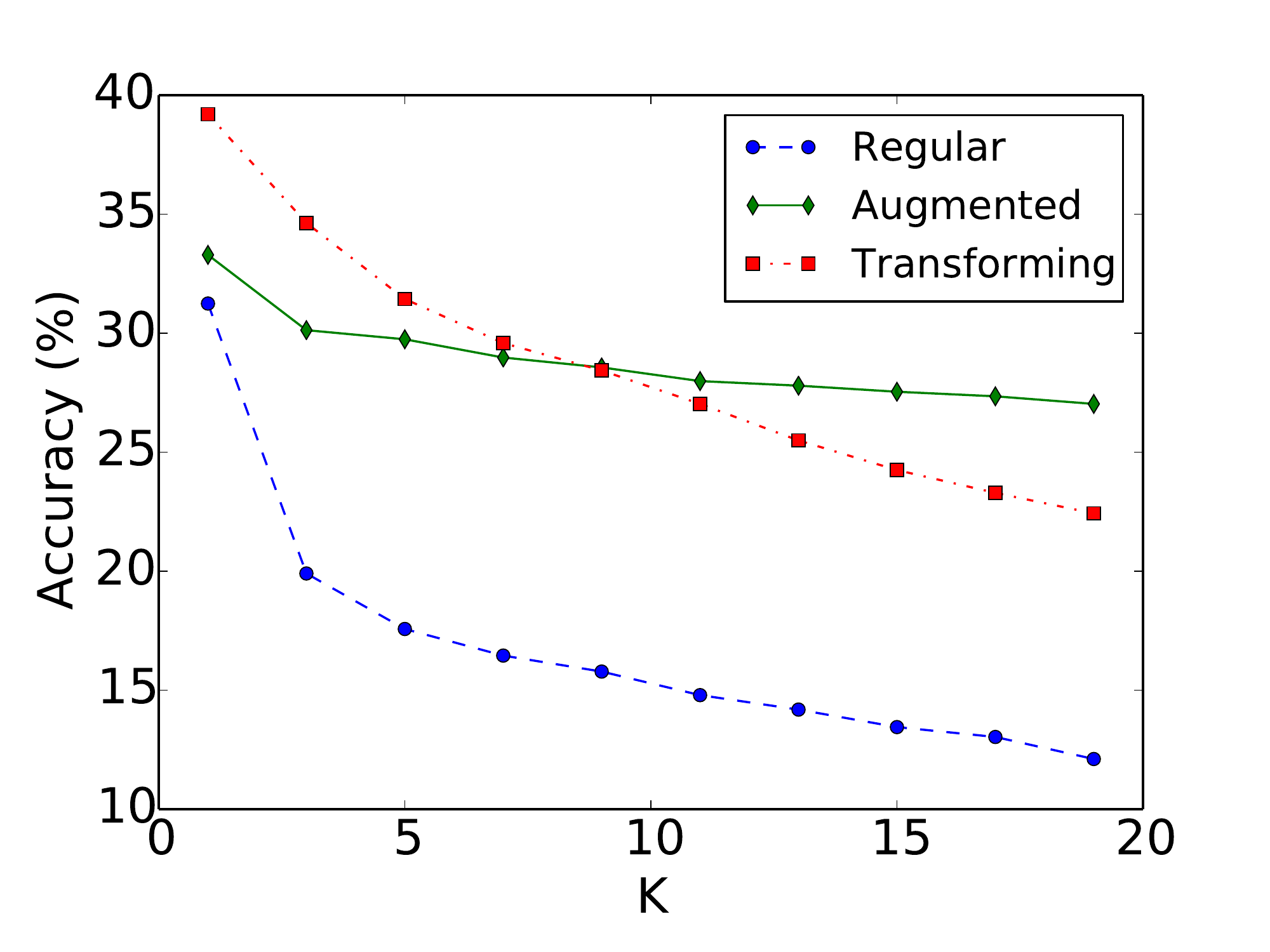}
\subcaption{TFDs1 with feature distance}    
\end{minipage}
\begin{minipage}[c]{0.32\linewidth}
\centering
\includegraphics[width=\figwidth]{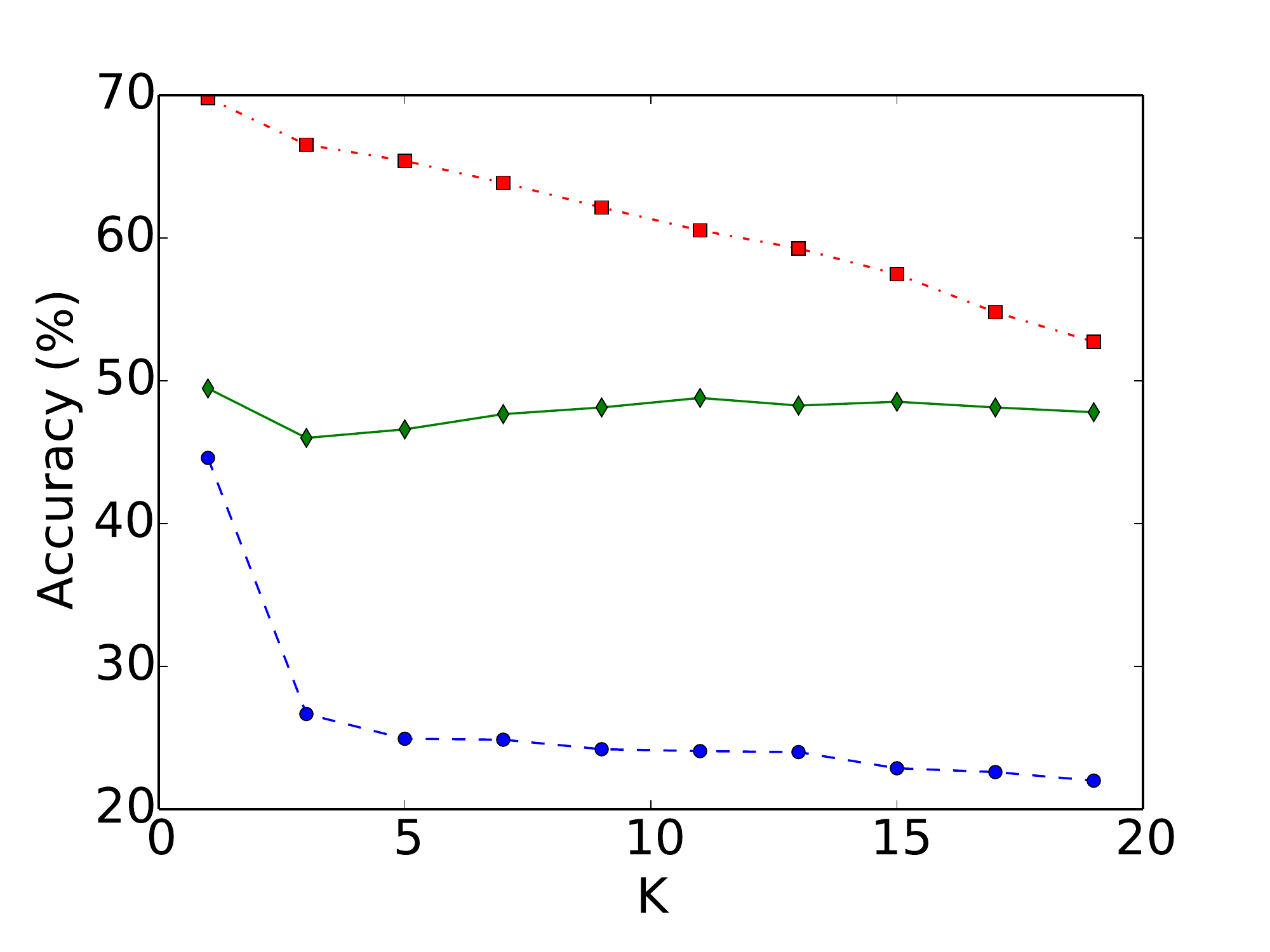}
\subcaption{TFDs2 with feature distance}    
\end{minipage}
\begin{minipage}[c]{0.32\linewidth}
\centering
\includegraphics[width=\figwidth]{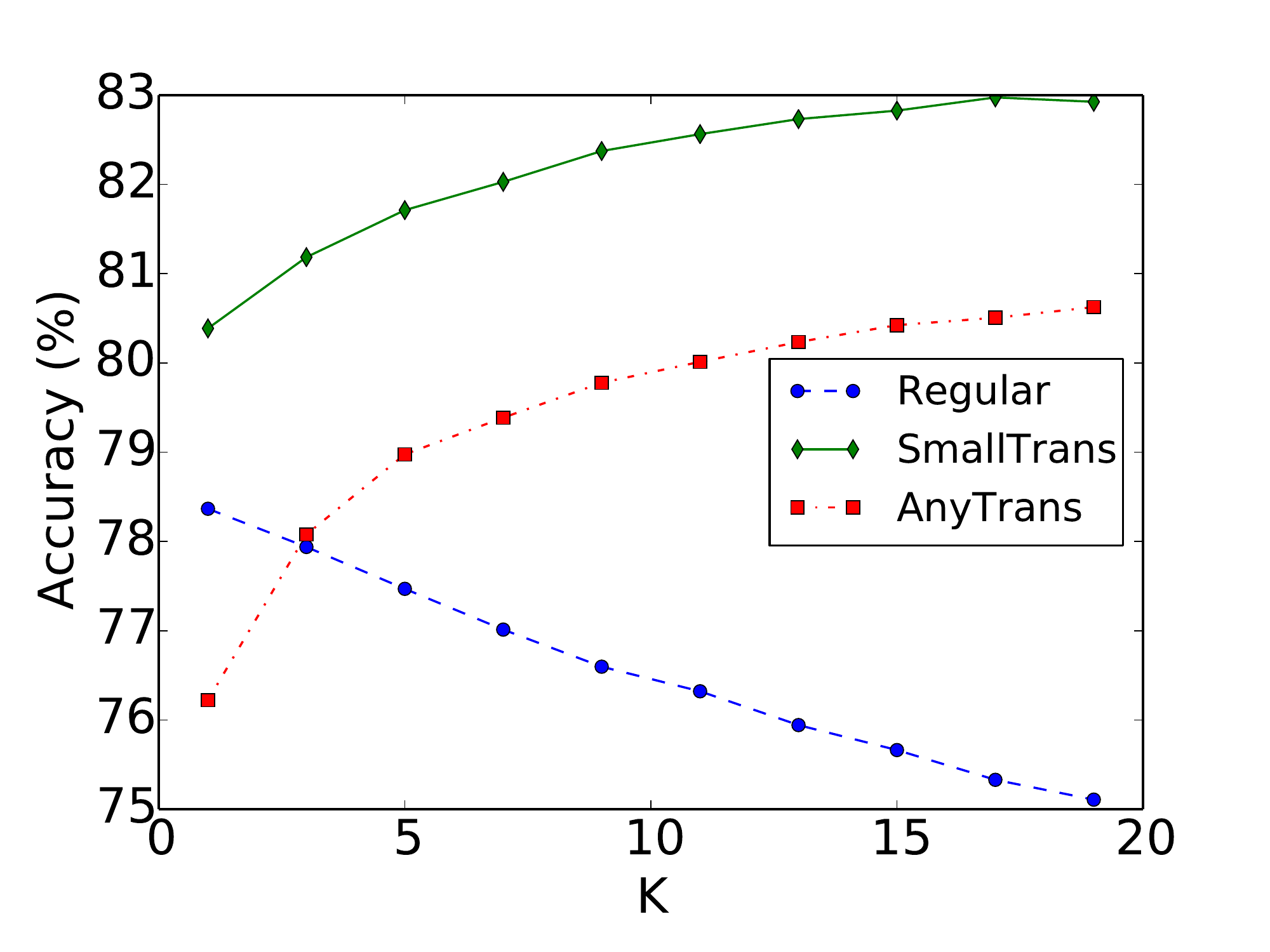}
\subcaption{NORB with pixel distance}
\end{minipage}

\caption{KNN accuracy vs.~$K$.
In (a) and (b), the dual-sided transforming KNN was performed and (b) shares the legend with (a). 
In (c), both SmallTrans and AnyTrans are single-sided.}
\label{fig: knn_k}

\vspace{-.4cm} 
\end{figure}

\subsubsection{KNN with a reduced database}
\label{sec: exp_knn_reduce}

\begin{figure}[t]
\centering
\begin{minipage}[c]{0.32\linewidth}
\centering
\includegraphics[width=\figwidth]{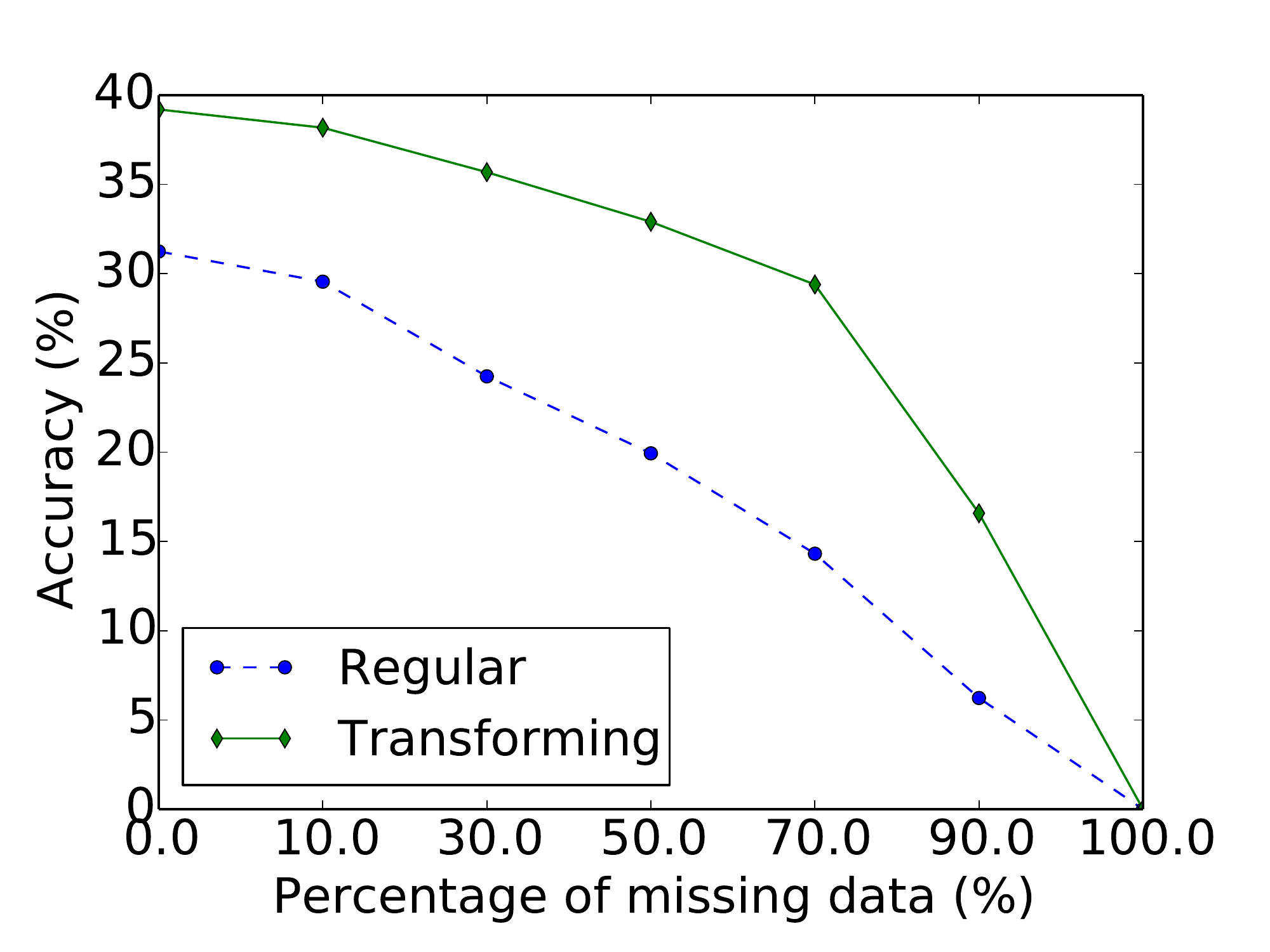}
\subcaption{TFDs1 with feature distance}    
\end{minipage}
\begin{minipage}[c]{0.32\linewidth}
\centering
\includegraphics[width=\figwidth]{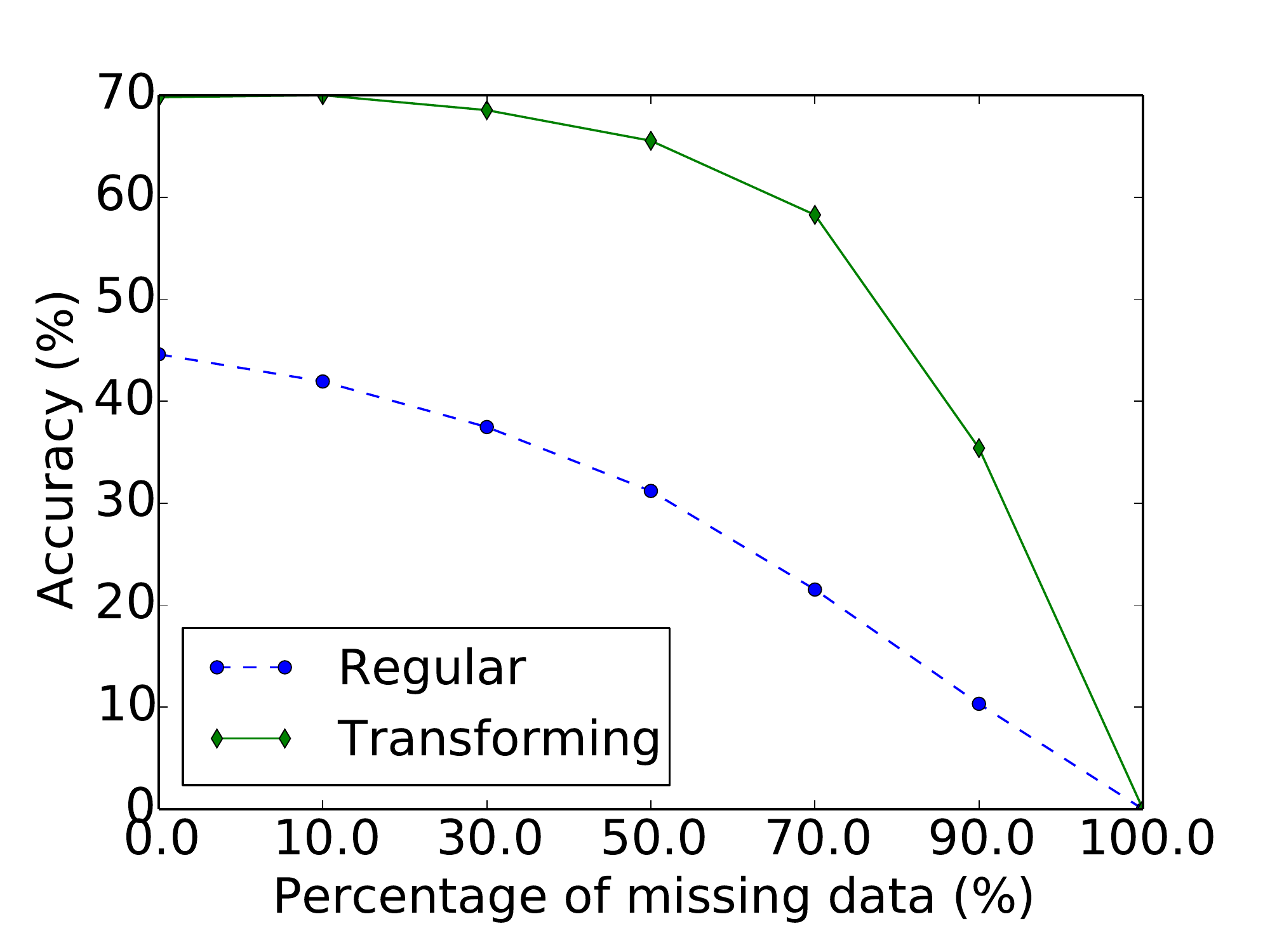}
\subcaption{TFDs2 with feature distance}    
\end{minipage}
\begin{minipage}[c]{0.32\linewidth}
\centering
\includegraphics[width=\figwidth]{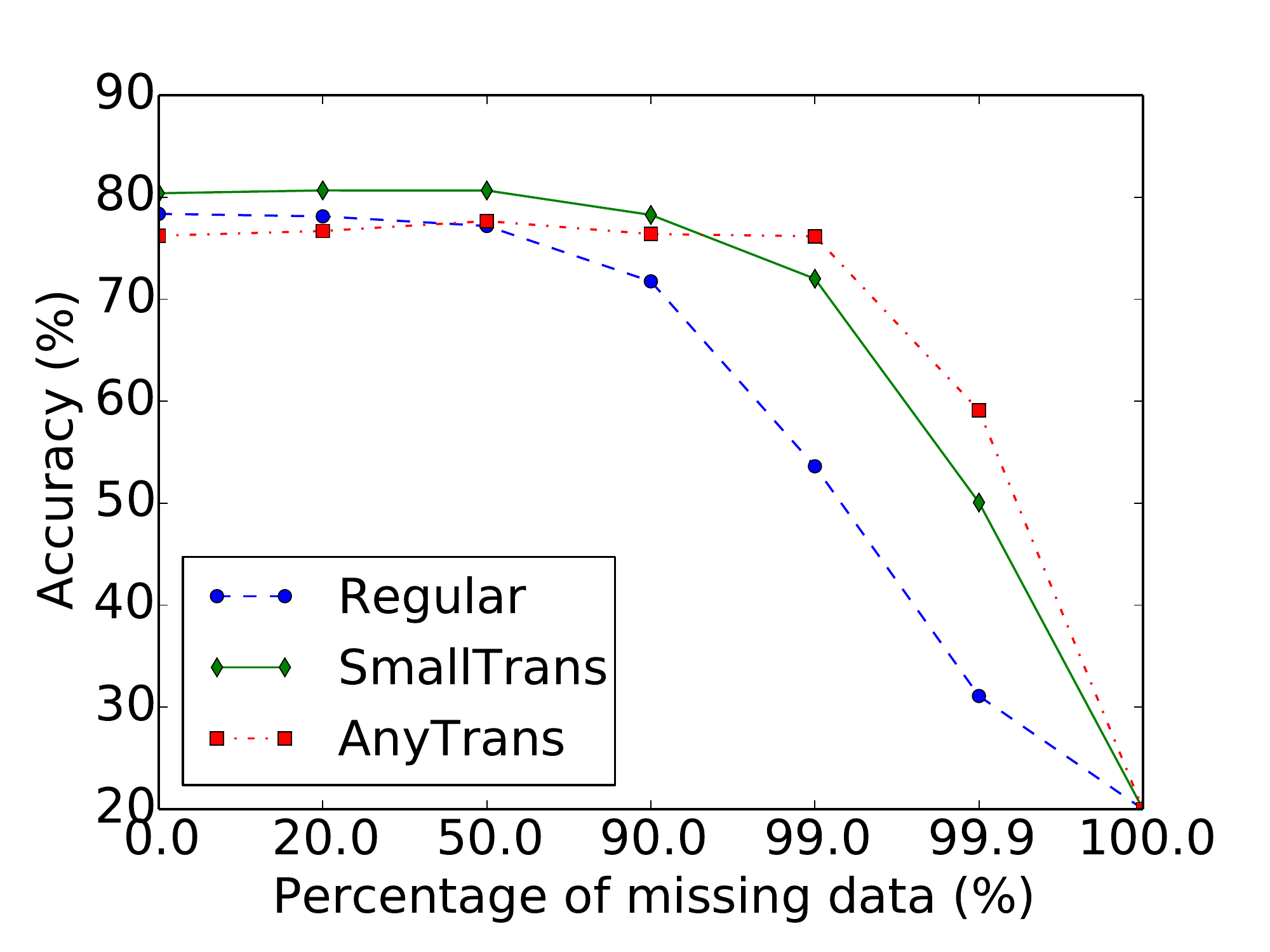}
\subcaption{NORB with pixel distance}    
\end{minipage}
\caption{KNN accuracy with reduced databases. In (a) and (b),
  dual-sided transforming KNN was performed. In (c), both SmallTrans
  and AnyTrans are single-sided. $K=1$ for all plots.}
\label{fig: knn_reduce}
\vspace{-.5cm} 
\end{figure}

As previously mentioned, under transforming KNN, each example in the
database can span a manifold and covers a ``volume''.  We examine this
assumption by performing KNN on reduced databases and showing the
relationship between accuracy and missing data rate (randomly
discarded) in Fig.~\ref{fig: knn_reduce}.  For both TFD and NORB,
transforming KNN works reasonably well with missing data.  Their
performance are still comparable to the regular KNN with complete
database, when 70\% (TFDs1), 80\% (TFDs2) and even 99\% (NORB,
AnyTrans) data are missing.

For TFD, missing data results in missing identities. Therefore, no
matter how good the relational model is, a performance drop is
inevitable.  In comparison, NORB contains only 5 classes.  This makes
it possible to maintain performance under a very high missing data
rate.  In Fig.~\ref{fig: knn_reduce} (c), AnyTrans KNN maintains its
accuracy even when 99\% data are missing (199 examples left in
database); SmallTrans KNN is still better than regular KNN when 90\%
data are missing.

SmallTrans KNN performs better than AnyTrans KNN when
the missing data rate is low, but degrades as the missing data rate
increases.  
We suspect that the fgRBM has difficulty encoding complicated
transformations (e.g.~arbitrary 3D changes of objects).
This conjecture (on the complexity of transformation) is 
supported by the fact that the best cross-validated fgRBM
dimensionality is 128 for SmallTrans and 256 for AnyTrans.

Although the AnyTrans fgRBM might have learned less realistic
transformations, it seems to be able to bridge large transformation
gaps when the database is sparse.  It maintains its
performance even when 99\% data are missing, and still provides a 60\%
accuracy when 99.9\% data are missing.
This hints at the potential of transforming KNN on ``weakly'' labeled
datasets, where only a small portion of data have class labels and the
majority of the data has weaker ``relational labels'' indicating image
pairs. Relational labels can be acquired from video.  Finally, using a
reduced KNN database could potentially speed up transforming KNN by
100 times at test time. This alleviates the additional computational
burden brought by test-time optimization.

\vspace{-0.15cm} 
\section{Conclusion}
\vspace{-0.18cm} 
The key novelty in our work is the idea of augmenting learned
similarity functions with latent variables capturing factors of
variation. Although we were inspired by the evidence of a mental
faculty for the simulation of spatial transformations, we believe that
this is just one example out of a richer class of dynamic similarity
models achievable within this framework. 
That is to say, the proposed framework is generic, since it is 
composed of interchangeable components.

Nevertheless, performing inference or optimization over latent
variables while comparing examples is much more computationally
demanding than the typical ``test-time'' application of learned
similarity models. This can be partially alleviated by a reduced need
for large databases. Unfortunately our method precludes the use of approximate nearest
neighbour techniques, which are typically used on large-scale problems
of the type we considered. We have, however, achieved modest
gains by parallelizing on GPUs. We believe that the computational cost
is the main concern in considering similarity models with latent
variables and we intend to address this issue in future work.

In each of the two datasets we considered, only a single
transformation class was learned. Relational models like the fgRBM and
relational autoencoder can, in theory, capture a rich set of
transformations. Therefore another avenue of future work is
demonstrating the efficacy of transforming similarity across a wider
range of transformations.






\bibliographystyle{plain}
\renewcommand{\refname}{\normalfont\normalsize\bfseries References}
\small{
\bibliography{transim}
}

\end{document}